\documentclass[runningheads]{llncs}
\usepackage{eccv}
\usepackage{eccvabbrv}
\usepackage{amsmath,amssymb,amsfonts}
\usepackage{multirow}
\usepackage{hhline} 
\usepackage{float}
\usepackage{dblfloatfix} 
\usepackage{graphicx}
\usepackage{booktabs}
\usepackage[accsupp]{axessibility}
\usepackage{hyperref}
\usepackage{orcidlink}
\newcommand{\scheme}{Chameleon}

\begin{document}

\title{Personalized Privacy Protection Mask Against Unauthorized Facial Recognition} 
\titlerunning{Personalized Privacy Protection Mask Against Unauthorized FR}

\author{Ka-Ho Chow\inst{1}\orcidlink{0000-0001-5917-2577} \and
Sihao Hu\inst{2}\orcidlink{0000-0003-3297-6991} \and
Tiansheng Huang\inst{2}\orcidlink{0000-0002-4557-1865} \and
Ling Liu\inst{2}\orcidlink{0000-0002-4138-3082}}

\authorrunning{K.-H. Chow et al.}

\institute{The University of Hong Kong\and
Georgia Institute of Technology\\
\email{kachow@cs.hku.hk, \{sihaohu,thuang374\}@gatech.edu, lingliu@cc.gatech.edu}}

\maketitle

\begin{abstract}
	Face recognition (FR) can be abused for privacy intrusion. Governments, private companies, or even individual attackers can collect facial images by web scraping to build an FR system identifying human faces without their consent. This paper introduces \scheme{}, which learns to generate a user-centric personalized privacy protection mask, coined as P3-Mask, to protect facial images against unauthorized FR with three salient features. First, we use a cross-image optimization to generate one P3-Mask for each user instead of tailoring facial perturbation for each facial image of a user. It enables efficient and instant protection even for users with limited computing resources. Second, we incorporate a perceptibility optimization to preserve the visual quality of the protected facial images.	Third, we strengthen the robustness of P3-Mask against unknown FR models by integrating focal diversity-optimized ensemble learning into the mask generation process. Extensive experiments on two benchmark datasets show that \scheme{} outperforms three state-of-the-art methods with instant protection and minimal degradation of image quality. Furthermore, \scheme{} enables cost-effective FR authorization using the P3-Mask as a personalized de-obfuscation key, and it demonstrates high resilience against adaptive adversaries. 
\end{abstract}

\section{Introduction}\label{sec:introduction}
Face recognition (FR) has long been investigated due to its potential for enhancing security and convenience in various domains~\cite{sahani2015web,pinto2011scaling,liu2018smart}. Many pretrained FR models are available online~\cite{insightface,huggingface}. Once a face database (a.k.a. the gallery) with facial images for each person of interest is provided, these pretrained models can be used to recognize them~\cite{wang2021deep}. 

\begin{figure} 
	\centering
	\includegraphics[width=0.88\linewidth]{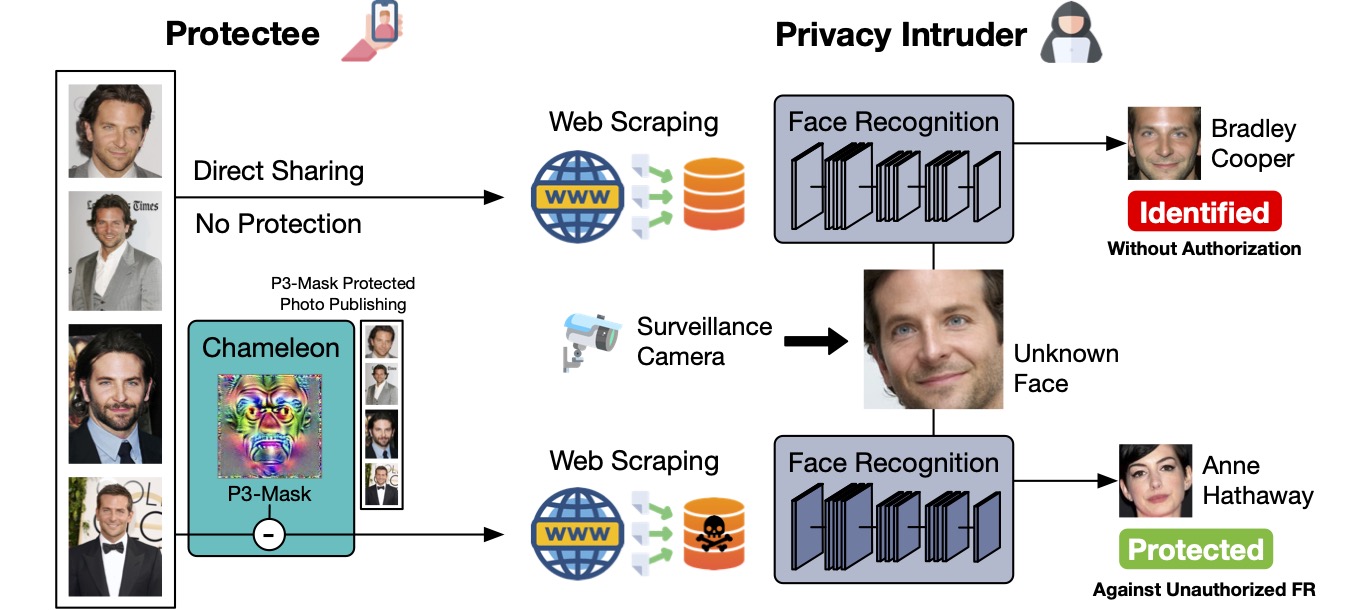}
	\caption{(Top) Without protection, the privacy intruder can build a face database by web scraping and identify the unknown face. (Bottom) \scheme{} learns the facial signature of the user (protectee) to generate a P3-Mask, which can be applied to protect any facial images before sharing them online against unauthorized FR.}
	\label{fig:intro}
\end{figure}

While empowering many life-enriching applications, FR can be abused to cause serious privacy issues~\cite{fr-pri-legal}. Privacy intruders can build a face database of victims of interest from publicly available facial images on the Internet by web scraping ({Figure~\ref{fig:intro}} (top)). 
By utilizing this database, the adversary can perform unauthorized FR of individual users for stalking victims~\cite{stalker}, intruding on victims' privacy by flooding targeted ads~\cite{retail}, or facilitating criminals to commit identity fraud~\cite{identity-fraud}. This is a real threat. For example, companies like Clearview~\cite{clearview} and PimEyes~\cite{pimeyes} have collected billions of online images and can recognize millions of citizens without their consent.

Such a facial privacy threat has motivated the development of anti-FR technologies~\cite{wenger2021sok}, which allow users (protectees) to preprocess their facial images before sharing them online. We argue that a competitive anti-FR solution should possess the following four properties. {First}, from the robustness perspective, the protected facial images should not be matched by an FR model to a query (probe) image with a correct identity (e.g., Bradley Cooper is recognized as Anne Hathaway in Figure~\ref{fig:intro} (bottom)), including those FR models possibly unknown during the preprocessing but are used by the privacy intruder. {Second}, from the efficiency perspective, the protection should be done instantly on any device with or without AI accelerators to maintain user experience and support equitable access to privacy protection. {Third}, from the perception usability perspective, the protected image should (i) preserve the visual quality rather than using excessive noise and (ii) be visually recognizable by humans to be the same person as the one in the original unprotected photo. {Finally}, the service usability for authorized FR providers is another important property. Existing approaches treat all FR models as unauthorized. We argue that anti-FR solutions should also allow users to grant FR permission to authorized models \emph{cost-effectively}. This user-initiated de-obfuscation capability is critical to those, e.g., who publish their photos on a social media platform with privacy protection and, at the same time, wish to authorize the platform to perform some FR services (e.g.,  face tagging~\cite{stone2008autotagging}) without sending and storing multiple versions of the same photo. 

Based on the above objectives, we introduce \scheme{}, a user-centric facial privacy protection system with three contributions. {First}, we develop an optimization algorithm to construct a \underline{P}ersonalized \underline{P}rivacy \underline{P}rotection mask, coined as P3-Mask, for each user. The P3-Mask of a user can be used to protect any facial images of the same user with minimal impact on image quality, including those facial images unseen during the P3-Mask generation process. {Second}, it boosts the robustness of P3-Mask against unknown FR models by a principled approach, leveraging ensemble learning with models of high focal diversity. The per-user P3-Mask provides higher robustness against unknown FR models while preserving perception usability and the service usability for authorized FR providers. \scheme{} users can utilize their own P3-Mask to obfuscate their facial images prior to public release, and the same P3-Mask can be used as a personalized de-obfuscation key to grant authorized FR service providers the ability to restore the facial signature of photos for correct recognition. {Third}, we conduct extensive experiments on two FR benchmark datasets~\cite{ng2014data,LFWTech} to analyze the shortcomings of state-of-the-art anti-FR solutions~\cite{shan2020fawkes,yang2021towards,zhong2022opom} and show that \scheme{} can better protect against unknown FR models with a high success rate. The protection is lightweight and in real-time. It remains effective under adaptive privacy intruders deploying various strategies to counter \scheme{}.

\section{Related Work}\label{sec:relatedwork}
Existing anti-FR approaches can be broadly classified into two categories. Synthesis-based approaches~\cite{sun2018hybrid,li2019anonymousnet,choi2018stargan,hu2022protecting,deb2020advfaces,zhu2019generating} use generative adversarial networks (GANs)~\cite{mirza2014conditional} to synthesize a face to replace the original one for protection. While the synthetic faces look realistic, they appear to be strangers, not recognizable even by the users themselves, failing to meet the perception usability requirement. 

\scheme{} falls into the second category, which applies small changes to facial images~\cite{chow2024diversity}. Fawkes~\cite{shan2020fawkes} formulates an untargeted attack to push the facial image away from the original location in the embedding space; TIP-IM~\cite{yang2021towards} improves the image quality with MMD~\cite{borgwardt2006integrating}; LowKey~\cite{cherepanova2021lowkey} improves the robustness under the image processing pipeline in an end-to-end ML system by incorporating it into the optimization process. These techniques can preserve the visual identity of protected faces. However, they both require iterative optimization for each image. Even for a GPU server, it can take over $100$ seconds to perturb one facial image (Section~\ref{sec:eval-cost}). In contrast, \scheme{} provides instant protection on any device, with protection even stronger than those spending minutes to find the best perturbation for each image. OPOM~\cite{zhong2022opom} attempts to improve efficiency by finding the embedding subspace enclosing facial images of the same person and generates a privacy protection mask to push any images away from it. However, it fails to maintain  protection effectiveness when image processing operations are applied to protected images (Section~\ref{sec:prot-effect}), and the mask degrades the image quality much more significantly than \scheme{} (Section~\ref{sec:eval-cost}). 

\section{Overview}\label{sec:overview}
To build an FR system with a pretrained model, the owner first collects a gallery dataset (face database) $\boldsymbol{\mathcal{D}}$, which includes the facial images of individuals of interest. Then, the pretrained FR model $F$ is used to map each gallery image $\tilde{\boldsymbol{x}}\in\boldsymbol{\mathcal{D}}$ to the embedding $F(\tilde{\boldsymbol{x}})$, and those embeddings should form clusters corresponding to different people. When a facial image $\boldsymbol{x}$ of an unknown identity, called the probe image, is given, the FR system can use the same FR model $F$ to map it to an embedding $F(\boldsymbol{x})$ and use the identity of the nearest gallery image to be the identity of the unknown person. The identity,  $\mathcal{FR}(\boldsymbol{x};F, \boldsymbol{\mathcal{D}})$, of the probe image $\boldsymbol{x}$ given the FR model $F$ and the gallery dataset $\boldsymbol{\mathcal{D}}$ is  defined as:
\begin{equation}\label{eq:standard-fr}\small
	\mathcal{FR}(\boldsymbol{x};F, \boldsymbol{\mathcal{D}})=\mathcal{I}\big(\underset{\tilde{\boldsymbol{x}}\in\boldsymbol{\mathcal{D}}}{\arg\min}\;\textsc{Dist}(F(\boldsymbol{x}),F(\tilde{\boldsymbol{x}}))\big),
\end{equation}
where $\mathcal{I}(\tilde{\boldsymbol{x}})$ denotes the identity of the gallery image $\tilde{\boldsymbol{x}}$, which is known to the FR system, and $\textsc{Dist}(\cdot,\cdot)$ is a distance function such as Euclidean distance. 

\textbf{Threat Model.} We consider a threat model where a privacy intruder conducts web scraping to collect images containing citizens' faces. Web scraping is large-scale and untargeted. Many citizens are included in the face database, and the privacy intruder may not know them. Given a probe image taken by, e.g., a stalker's camera, the privacy intruder uses an FR model on the face database to search for matched image(s). Then, the privacy intruder can analyze the associated metadata, such as the web page from where it was downloaded or even the exact identity. The goal of \scheme{} is to allow the user to preprocess her facial images before posting them online. Even if scraped and included in the face database, they will not be matched to a probe image of her. 

\subsection{\scheme{} Design}
\begin{figure}[t]
	\centering
	\includegraphics[width=\textwidth]{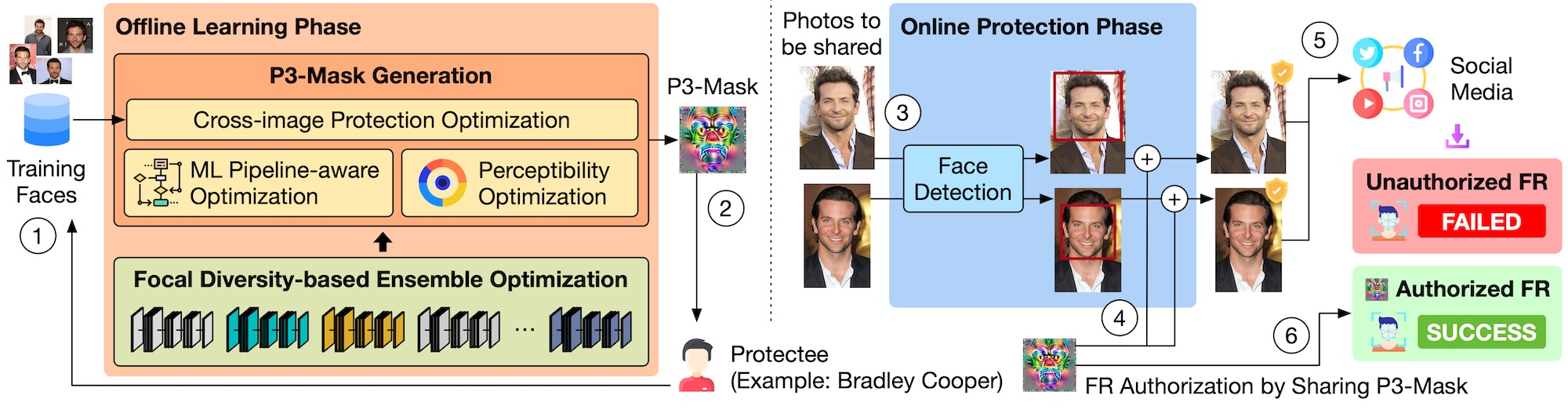}
	\caption{An overview of \scheme{}'s two-phase workflow for offline learning to generate P3-Mask  and online protection with P3-Mask for personalized facial signature masking. The P3-Mask can protect \emph{any} facial images of the same person without further learning.}\label{fig:overview}
\end{figure}

\scheme{}  performs preprocessing of the user's photos before sharing them online by applying the user-specific P3-Mask. This  mask is generated offline for a \scheme{} user. {Figure~\ref{fig:overview}} provides an overview of \scheme{}'s workflow.

\textbf{Offline Learning Phase.} \scheme{} learns the unique facial signature of a user from a few facial images of her using the P3-Mask generator (Section~\ref{sec:generation}), as shown in {Figure~\ref{fig:overview}} \textcircled{1}. P3-Mask is designed to apply directly to any facial images of the same user, including those unseen during offline learning, and is robust against lossy image processing operations in an end-to-end ML system without compromising the image quality. We use a focal diversity-optimized ensemble  learning method to find a team of FR models such that the P3-Mask generated against them has strong robustness and is effective in countering other FR models that are unknown during offline learning.

\textbf{Online Protection Phase.} After the offline learning, \textcircled{2} the P3-Mask can be sent to the user's local device (e.g., the mobile phone). For a photo of the user to be protected,  \textcircled{3} a lightweight face detection model, such as MediaPipe~\cite{lugaresi2019mediapipe}, can be used by the \scheme{} user running on the user's device to locate the face region and \textcircled{4} instantly protect it with the P3-Mask before the user shares it online \textcircled{5}. The user can also authorize some trusted entities to conduct FR on their shared photos protected with P3-Mask. The user-specific de-obfuscation key \textcircled{6} allows the protected photos to be restored for authorized FR.

\section{P3-Mask Generation}\label{sec:generation}
For a user $\mathcal{P}$, \scheme{} generates a P3-Mask by learning the facial signature $\boldsymbol{\mathcal{M}}_{{\mathcal{P}}}$ from a set of facial images the user provides. Several key factors need to be accomplished. 
(1) We need to promote cross-image protection as an optimization objective to generate the most representative facial signature applicable to protect any facial images of the user $\mathcal{P}$, including those unknown ones during the generation process (Section~\ref{sec:cipo}). 
(2) We need to control the amount of perturbation introduced by the P3-Mask. When applied to an unprotected image by removing the learned facial signature, it ensures minimal perception loss to preserve the visual quality (Section~\ref{sec:io}). 
(3) We need to keep the protected image far away from the original image in the embedding space. Multiple FR models should be used to generate alternative embedding representations to enhance generalizability against unknown models (Section~\ref{sec:teaming}).

\subsection{Cross-image Protection Optimization}\label{sec:cipo}
The cross-image protection capability comes from iterative learning on a set of training images the user provides. The idea is to keep fine-tuning the P3-Mask so that it can offer protection simultaneously to training images while preserving image quality. At the $t$-th iteration, it samples a mini-batch $\boldsymbol{\mathcal{B}}$ from the training dataset $\boldsymbol{\Omega}$ containing photos of user $\mathcal{P}$. Then, we find the modification to the P3-Mask that can lead to better protection on each image in the mini-batch with better visual quality with the following operations:
\begin{equation}\label{eq:cham}\small
	\boldsymbol{\mathcal{M}}_\mathcal{P}^{t+1}=\textsc{ Clip}_{[-\epsilon,\epsilon]}(\boldsymbol{\mathcal{M}}^{t}_{\mathcal{P}}-\eta\textsc{Sign}(\frac{1}{\vert\boldsymbol{\mathcal{B}}\vert}\sum_{\mathcal{X}\in\boldsymbol{\mathcal{B}}}\nabla_{\boldsymbol{\mathcal{M}}^t_\mathcal{P}}\mathcal{L}(\mathcal{X},\boldsymbol{\mathcal{M}}^{t}_{\mathcal{P}};\boldsymbol{\mathcal{T}}, \omega))).
\end{equation}
Specifically, for each image $\mathcal{X}\in\boldsymbol{\mathcal{B}}$, we use the P3-Mask learned up to the current iteration, $\boldsymbol{\mathcal{M}}^t_\mathcal{P}$, to protect it and compute the loss designed with dual goals:
\begin{equation}\label{eq:L}\small
	\mathcal{L}(\mathcal{X},\boldsymbol{\mathcal{M}}^{t}_{\mathcal{P}};\boldsymbol{\mathcal{T}}, \omega)=\mathcal{L}_{\text{Protect}}(\mathcal{X},\boldsymbol{\mathcal{M}}^{t}_{\mathcal{P}};\boldsymbol{\mathcal{T}})+
	\mathcal{L}_{\text{Percept}}(\mathcal{X},\boldsymbol{\mathcal{M}}^{t}_{\mathcal{P}};\omega).
\end{equation}
It uses $\mathcal{L}_{\text{Protect}}$ to learn how to adjust the P3-Mask $\boldsymbol{\mathcal{M}}^{t}_{\mathcal{P}}$ from the previous iteration to better protect against a team of pre-selected FR models $\boldsymbol{\mathcal{T}}$ and $\mathcal{L}_{\text{Percept}}$ to preserve the image quality controlled by $\omega$. We use the \textsc{Sign} of the gradients to update the P3-Mask with a learning rate $\eta$. A clipping function $\textsc{Clip}_{[-\epsilon,\epsilon]}$ is applied to ensure the changes made by the P3-Mask on the facial image are bounded by $\epsilon$. Such cross-image protection is not limited to those in the training dataset $\boldsymbol{\Omega}$ but also unseen images of the same user. We next discuss the design of $\mathcal{L}_{\text{Protect}}$, and $\mathcal{L}_{\text{Percept}}$ will be detailed in Section~\ref{sec:io}.

\begin{figure}[t]
	\centering
	\includegraphics[width=0.95\textwidth]{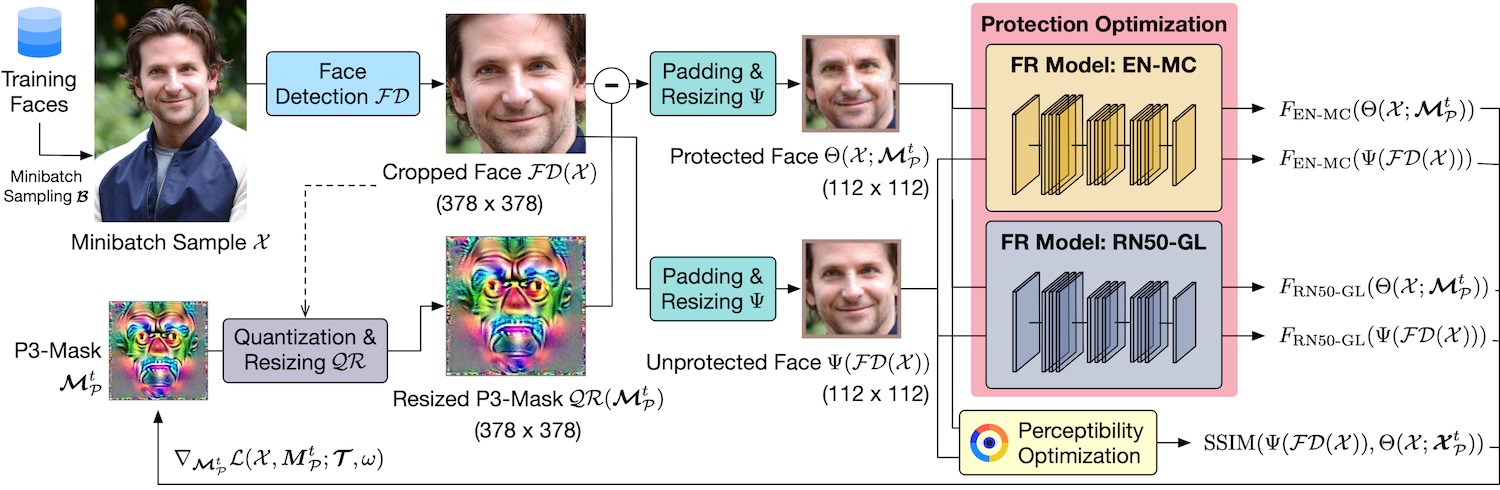}
	\caption{The iterative generation of P3-Mask for a user. \scheme{} goes through multiple facial images of the same user to optimize a P3-Mask with ML pipeline awareness.}\label{fig:mask-generation}
\end{figure}

For privacy protection, P3-Mask maximizes the distance between the protected and unprotected images in the embedding spaces of a team $\boldsymbol{\mathcal{T}}$ of pre-selected FR models. We incorporate image processing operations into the optimization to avoid those lossy operations degrading P3-Mask's protection. Consider the raw training image $\mathcal{X}$ in the mini-batch in {Figure~\ref{fig:mask-generation}}. 
We first conduct face detection to produce the cropped face $\mathcal{FD}(\mathcal{X})$. Quantization and resizing are executed on the P3-Mask to match the resolution of the cropped face, denoted by $\mathcal{QR}(\boldsymbol{\mathcal{M}}^t_\mathcal{P})$. Then, we can apply the mask to the face for protection:
\begin{equation}\small
\Theta(\mathcal{X}; \boldsymbol{\mathcal{M}}^t_{\mathcal{P}})
=\Psi(\textsc{Clip}_{[0,255]}(\mathcal{FD}(\mathcal{X})-\mathcal{QR}(\boldsymbol{\mathcal{M}}^t_\mathcal{P}))),
\end{equation}
where $\Psi$ is a padding and resizing function to meet the input resolution requirement of an FR model (e.g., $(112\times112)$ in ArcFace~\cite{deng2019arcface}). Note that to simplify the notation, we removed certain arguments without causing confusion.

The protection optimization in P3-Mask maximizes the average arccosine distance~\cite{deng2019arcface}, $\textsc{ArcCos}$, between the embedding of the protected face $F(\Theta(\mathcal{X}; \boldsymbol{\mathcal{M}}^t_{\mathcal{P}}))$ and the embedding of its unprotected counterpart $F(\Psi(\mathcal{FD}(\mathcal{X})))$ extracted by every FR model $F\in\boldsymbol{\mathcal{T}}$ in the team:
\begin{equation}\label{eq:prot}\small
	\mathcal{L}_{\text{Protect}}(\mathcal{X},\boldsymbol{\mathcal{M}}^{t}_{\mathcal{P}};\boldsymbol{\mathcal{T}})
	=\frac{-1}{\vert\boldsymbol{\mathcal{T}}\vert}\sum_{F\in\boldsymbol{\mathcal{T}}}\textsc{ArcCos}(F(\Psi(\mathcal{FD}(\mathcal{X}))), F(\Theta(\mathcal{X}; \boldsymbol{\mathcal{M}}^t_{\mathcal{P}}))).
\end{equation}

\subsection{Perceptibility Optimization}\label{sec:io}
Preserving the visual quality of the facial image is necessary to maintain the usability of the protected image in practice. Hence, we incorporate a perceptibility optimization term in Equation~\ref{eq:L}. The idea is to minimize the perceptual difference between unprotected and protected images while learning a user's facial signature, such that removing the facial signature from a given facial image will have minimal impact on the visual quality of the protected version of the original image. We use the structural similarity (SSIM)~\cite{wang2004image} to capture perceptual differences, as it has been shown to align with the human perception system:
\begin{equation}\label{eq:l-imp}\small
	\mathcal{L}_{\text{Percept}}(\mathcal{X},\boldsymbol{\mathcal{M}}^{t}_{\mathcal{P}};\omega)=\lambda_{\text{SSIM}}\max\bigg[
	\frac{1-\textsc{SSIM}(\Psi(\mathcal{FD}(\mathcal{X})), \Theta(\mathcal{X};\boldsymbol{\mathcal{M}}^t_\mathcal{P}))}{2}-\omega, 0\bigg].
\end{equation}
The parameters, $\omega$ and $\lambda_{\text{SSIM}}$, enable two features. $\omega$ controls the SSIM degradation. Any SSIM degradation greater than $2\omega$ will cause the term to be non-zero, making the optimization adjust the P3-Mask reducing its impact on image quality. $\lambda_{\text{SSIM}}$ balances the importance of privacy protection and perceptibility. We use dynamic scheduling~\cite{shan2020fawkes} such that it will be adjusted automatically. 

\subsection{Focal Diversity Ensemble Optimization}\label{sec:teaming}
\begin{figure}[t]
\begin{minipage}{.41\textwidth}
\centering
\begin{subfigure}{\linewidth}
	\centering
	\includegraphics[width=0.85\textwidth]{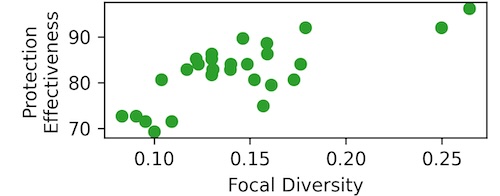}
	\caption{Two-model: $28$ Options}
	\label{fig:focal-motivation-2}
\end{subfigure}\vspace{0.5em}
\begin{subfigure}{\linewidth}
	\centering
	\includegraphics[width=0.85\textwidth]{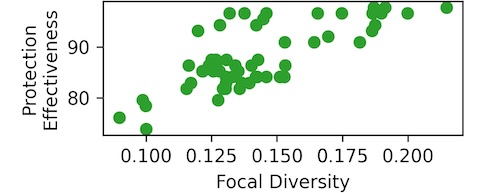}
	\caption{Three-model: $56$ Options}
	\label{fig:focal-motivation-3}
\end{subfigure}
\caption{Focal diversity provides a strong indicator of protection effectiveness for selecting an ensemble.}
\label{fig:focal-motivation}
\end{minipage}
\hfill
\begin{minipage}{0.57\textwidth}
	\centering
	\includegraphics[width=0.9\linewidth]{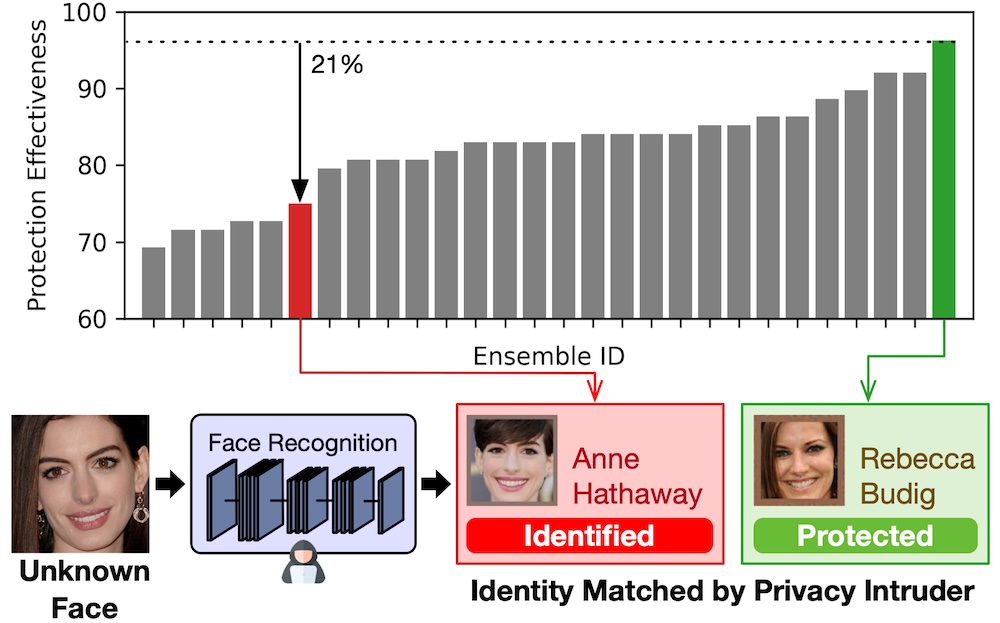}
	\caption{Out of $28$ options of two-model ensembles from a pool of eight models, the ensemble selected by our approach (green) leads to much better protection than a randomly selected one (red).}\label{fig:teaming-gap}
\end{minipage}
\end{figure}

For each original image and its protected version by removing the facial signature learned up to the current iteration, we use two FR models in  Figure~\ref{fig:mask-generation} to generate two  embeddings, which serve as alternative channels to learn a high-quality facial signature. Choosing FR models that best complement each other can achieve better protection than randomly selected FR models. We use an ensemble optimization method to select the best $k$-model ensemble among a pool of $N$ FR models. The key idea is to find an ensemble with members making decorrelated mistakes. To quantify this behavior, we leverage the focal diversity framework~\cite{wu2021boosting,chow2022boosting,chow2021robust}. As shown in {Figure~\ref{fig:focal-motivation}}, the higher the focal diversity of the ensemble (a green dot), the stronger the P3-Mask will be in protecting against other FR models \emph{not} used during the P3-Mask generation process.

Given a collection of $N$ FR models $\{F_1,...,F_N\}$, we first identify the negative samples for each FR model $F$ by locating validation images that $F$ fails to recognize their true identity. 
To rank ensembles of size $S$, we enumerate all ${N\choose S}$ combinations. For each combination (ensemble) $\boldsymbol{\mathcal{T}}$, we consider each member to be the focal model $F_{\text{focal}}$ and use its negative samples to statistically estimate the level of negative correlation $\lambda_{\text{focal}}(\boldsymbol{\mathcal{T}};F_{\text{focal}})$ between $F_{\text{focal}}$ and the remaining models in $\boldsymbol{\mathcal{T}}$, computed by measuring the degree of disagreements using the generalized non-pairwise measure~\cite{partridge1997software}. The same procedure is repeated by considering each member in the ensemble as the focal model, and the focal diversity of the ensemble $\boldsymbol{\mathcal{T}}$ is finalized as $d_{\text{focal}}(\boldsymbol{\mathcal{T}})=\frac{1}{S}\sum_{F_{\text{focal}}\in{\boldsymbol{\mathcal{T}}}}[1-\lambda_{\text{focal}}(\boldsymbol{\mathcal{T}}; F_{\text{focal}})]$.
Given a team size $S$, the ensemble with the highest focal diversity can be deployed. As shown in {Figure~\ref{fig:teaming-gap}}, the selected two-model team (green bar) is indeed the one leading to the strongest protection among all $28$ options, which is over $20\%$ stronger than a randomly selected ensemble (red bar). Note that we only need to analyze the failures of individual FR models, which is significantly more efficient than generating P3-Mask for each option and conducting evaluation.

\section{Online Image Protection and Refinement}
The user $\mathcal{P}$ obtains her P3-Mask $\boldsymbol{\mathcal{M}}_\mathcal{P}$ from \scheme{}. Given an image $\mathcal{X}$, she can obfuscate her facial identity on her device by applying the P3-Mask:
\begin{equation}\label{eq:encryption}\small
\textsc{Mask}(\mathcal{X};\boldsymbol{\mathcal{M}}_\mathcal{P})=\textsc{Clip}_{[0,255]}(\mathcal{FD}(\mathcal{X})-\mathcal{QR}(\boldsymbol{\mathcal{M}}_\mathcal{P})),
\end{equation}
which is the protected facial region to be put in the original image $\mathcal{X}$ to produce the protected version $\mathcal{X}'$ for sharing.

\scheme{} supports authorizing trusted third parties to conduct FR on a user's facial images in a space-time efficient manner without transmitting and storing multiple versions of the same photo (one protected for public sharing and one unprotected for internal FR).
By sharing the P3-Mask $\boldsymbol{\mathcal{M}}_\mathcal{P}$  as the key, the third parties can de-obfuscate the facial signature by unmasking:
\begin{equation}\label{eq:decryption}\small
\hspace*{-0.5em}\textsc{Unmask}(\mathcal{X}';\boldsymbol{\mathcal{M}}_\mathcal{P})=\textsc{Clip}_{[0, 255]}(\mathcal{FD}(\mathcal{X'})+\mathcal{QR}(\boldsymbol{\mathcal{M}}_\mathcal{P})).
\end{equation}
In Section~\ref{sec:decrypt-results}, we will show two properties of this process: 
(i) the signature-restored photos can be used for FR with no accuracy degradation, and
(ii) the protection can only be done by the P3-Mask of the same person in the facial image, and the restoration is successful only if the same key is used.

\section{Experimental Evaluation}\label{sec:evaluation}
We conduct experiments on FaceScrub~\cite{ng2014data} and LFW~\cite{LFWTech}. 
To provide a detailed analysis, ten celebrities are randomly selected as users on FaceScrub ({Table~\ref{tab:dataset}}). 
\begin{table}\scriptsize
	\begin{minipage}[t]{.485\textwidth}
		\centering 
		\caption{Ten example celebrities (users) on FaceScrub for analysis.}
		\label{tab:dataset}
		\begin{tabular}{|lcccc|}
			\hline
			\multicolumn{1}{|c|}{\multirow{2}{*}{\textbf{Name}}} &
			\multicolumn{1}{c|}{\multirow{2}{*}{\textbf{\begin{tabular}[c]{@{}c@{}}Probe\\ Images\end{tabular}}}} &
			\multicolumn{3}{c|}{\textbf{Gallery Images}} \\ \cline{3-5} 
			\multicolumn{1}{|c|}{} &
			\multicolumn{1}{c|}{} &
			\multicolumn{1}{l|}{\textbf{Seen}} &
			\multicolumn{1}{l|}{\textbf{Unseen}} &
			\multicolumn{1}{l|}{\textbf{Total}} \\ \hline
			\multicolumn{1}{|l|}{Morena Baccarin}      & \multicolumn{1}{c|}{11} & \multicolumn{1}{c|}{82}  & \multicolumn{1}{c|}{23} & 105 \\ \hline
			\multicolumn{1}{|l|}{Bradley Cooper}       & \multicolumn{1}{c|}{12} & \multicolumn{1}{c|}{89}  & \multicolumn{1}{c|}{25} & 114 \\ \hline
			\multicolumn{1}{|l|}{America Ferrera}      & \multicolumn{1}{c|}{16} & \multicolumn{1}{c|}{113} & \multicolumn{1}{c|}{32} & 145 \\ \hline
			\multicolumn{1}{|l|}{Gerard Butler}        & \multicolumn{1}{c|}{11} & \multicolumn{1}{c|}{82}  & \multicolumn{1}{c|}{24} & 106 \\ \hline
			\multicolumn{1}{|l|}{Eva Longoria}         & \multicolumn{1}{c|}{12} & \multicolumn{1}{c|}{91}  & \multicolumn{1}{c|}{26} & 117 \\ \hline
			\multicolumn{1}{|l|}{Melissa Egan} & \multicolumn{1}{c|}{12} & \multicolumn{1}{c|}{87}  & \multicolumn{1}{c|}{24} & 111 \\ \hline
			\multicolumn{1}{|l|}{Kim Cattrall}         & \multicolumn{1}{c|}{13} & \multicolumn{1}{c|}{96}  & \multicolumn{1}{c|}{27} & 123 \\ \hline
			\multicolumn{1}{|l|}{Allison Janney}       & \multicolumn{1}{c|}{12} & \multicolumn{1}{c|}{89}  & \multicolumn{1}{c|}{26} & 115 \\ \hline
			\multicolumn{1}{|l|}{Roma Downey}          & \multicolumn{1}{c|}{11} & \multicolumn{1}{c|}{78}  & \multicolumn{1}{c|}{22} & 100 \\ \hline
			\multicolumn{1}{|l|}{Steve Carell}         & \multicolumn{1}{c|}{13} & \multicolumn{1}{c|}{96}  & \multicolumn{1}{c|}{28} & 124 \\ \hline
			\multicolumn{1}{|l|}{Others (Noise)}        & \multicolumn{1}{c|}{/}  & \multicolumn{1}{c|}{/}   & \multicolumn{1}{c|}{/}  & 48368   \\ \hhline{|=|=|=|=|=|}
			\multicolumn{4}{|r|}{\textbf{Total Number of Gallery Images:}}                                                                                       & 49528   \\ \hline
		\end{tabular}
	\end{minipage}
	\hfill
	\begin{minipage}[t]{0.495\textwidth}
		\centering
		\caption{The collection of publicly available pretrained FR models.}
		\label{tab:model-collection}
		\begin{tabular}{|c|c|c|c|}
			\hline
			\multirow{2}{*}{\textbf{\begin{tabular}[c]{@{}c@{}}Model\\ ID\end{tabular}}} &
			\multirow{2}{*}{\textbf{\begin{tabular}[c]{@{}c@{}}Neural\\ Arch.\end{tabular}}} &
			\multirow{2}{*}{\textbf{\begin{tabular}[c]{@{}c@{}}Training\\ Dataset\end{tabular}}} &
			\multirow{2}{*}{\textbf{\begin{tabular}[c]{@{}c@{}}FR\\ Acc.\end{tabular}}} \\ 
			&              &             &   \\ \hline
			EN-MC    & EfficientNet & MS-Celeb-1M & \multicolumn{1}{c|}{94.94\%}          \\ \hline
			RN50-GL  & ResNet50     & Glint360K   & \multicolumn{1}{c|}{96.68\%}          \\ \hline
			RN50-MC  & ResNet50     & MS-Celeb-1M & \multicolumn{1}{c|}{89.25\%}          \\ \hline
			RN50-VF  & ResNet50     & VGG-Face2   & \multicolumn{1}{c|}{94.30\%}          \\ \hline
			RN50-WF  & ResNet50     & WebFace600K & \multicolumn{1}{c|}{96.72\%}          \\ \hline
			RN18-MC  & ResNet18     & MS-Celeb-1M & \multicolumn{1}{c|}{82.64\%}          \\ \hline
			RN34-MC  & ResNet34     & MS-Celeb-1M & \multicolumn{1}{c|}{84.49\%}          \\ \hline
			RN100-MC & ResNet100    & MS-Celeb-1M & \multicolumn{1}{c|}{91.85\%}          \\ \hline
		\end{tabular}
	\end{minipage}
\end{table}
For each user, we split their facial images into three parts: (i) $10\%$ are used as probe images 
($2$nd column), (ii) $70\%$ are used as gallery images 
and are used by \scheme{} to train the mask ($3$rd column), and (iii) $20\%$ are also used as gallery images but unseen during the training process ($4$th column). The last part is crucial to evaluate the protection of unseen images of the user. In total, we have $123$ facial images with ``unknown identities.'' Following relevant works~\cite{shan2020fawkes,zhong2022opom}, all facial images of other people are included in the gallery dataset as noise. The same splitting method is also used for LFW. By default, the results reported in this paper focus on FaceScrub due to similar observations on LFW. The source code is available at \url{https://github.com/git-disl/Chameleon}.

Public clouds (e.g., Azure) now require manual approval to use their FR services. Hence, we believe that privacy intruders will opt for deploying pretrained FR models, as many high-quality ones are available on the Internet and can be used out of the box. Due to the large number of possible FR algorithms and architectures, we first focus on the eight FR models listed in {Table~\ref{tab:model-collection}}, which are based on ArcFace~\cite{deng2019arcface}, the state-of-the-art FR algorithm, with varying neural architectures and training datasets. 
\scheme{} can be easily extended to incorporate any FR models by simply adding them to the collection. Nonetheless, in Section~\ref{sec:eval-time}, we will show that thanks to the focal diversity-optimized teaming, the P3-Mask generated from a collection of ArcFace-only models can also be effective in protecting against privacy intruders using models of other FR algorithms, such as FaceNet~\cite{schroff2015facenet} and MagFace~\cite{meng2021magface}. By default, we use the two-model team (EN-MC, RN50-GL) with the highest focal diversity (Section~\ref{sec:teaming}). The P3-Mask is trained for $50$ epochs on NVIDIA RTX 2080 SUPER GPU with $\eta=0.001$, $\vert\boldsymbol{\mathcal{B}}\vert=4$, $\epsilon=0.063$, and $\omega=0.03$. 
We provide details for reproducibility and additional results in the appendix.
\begin{table}[b]\scriptsize
\centering
\setlength\tabcolsep{1pt}
\caption{The P3-Mask (rescaled) for seven users on FaceScrub. \scheme{} learns the distinct facial signature for each of them. }
\label{tab:masks}
\begin{tabular}{|c|c|c|c|c|c|c|}
	\hline
	\textbf{M. Baccarin} & \textbf{B. Cooper}  & \textbf{A. Ferrera} & \textbf{G. Butler} &\textbf{E. Longoria}  & \textbf{M. Egan} &\textbf{K. Cattrall}      \\ \hline
	\includegraphics[width=47px]{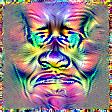} &    \includegraphics[width=47px]{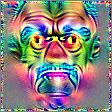}               & \includegraphics[width=47px]{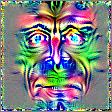}              & \includegraphics[width=47px]{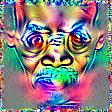}                 & \includegraphics[width=47px]{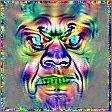}         & \includegraphics[width=47px]{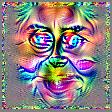} & \includegraphics[width=47px]{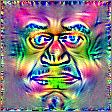}        \\ \hline
\end{tabular}
\end{table}

\subsection{Protection Effectiveness}\label{sec:prot-effect}
{Table~\ref{tab:masks}} shows the personalized mask for seven users on FaceScrub (other users are available in the appendix). Different users have distinct facial signatures (P3-Mask) learned by \scheme{}. 
We apply these masks to their respective gallery images and test the FR accuracy using probe images. All probe images can be correctly identified when no protection mechanism is used. 
Under \scheme{}, the probe images should be matched to an incorrect identity, resulting in a low FR accuracy. Hence, we define an evaluation metric, \underline{P}rotection \underline{S}uccess \underline{R}ate (PSR), to be $(100-\textsc{FR Accuacy})$ reported in percentages. 

\begin{figure}[t]
	\centering
	\begin{subfigure}{0.49\linewidth}
		\centering
		\includegraphics[width=\linewidth]{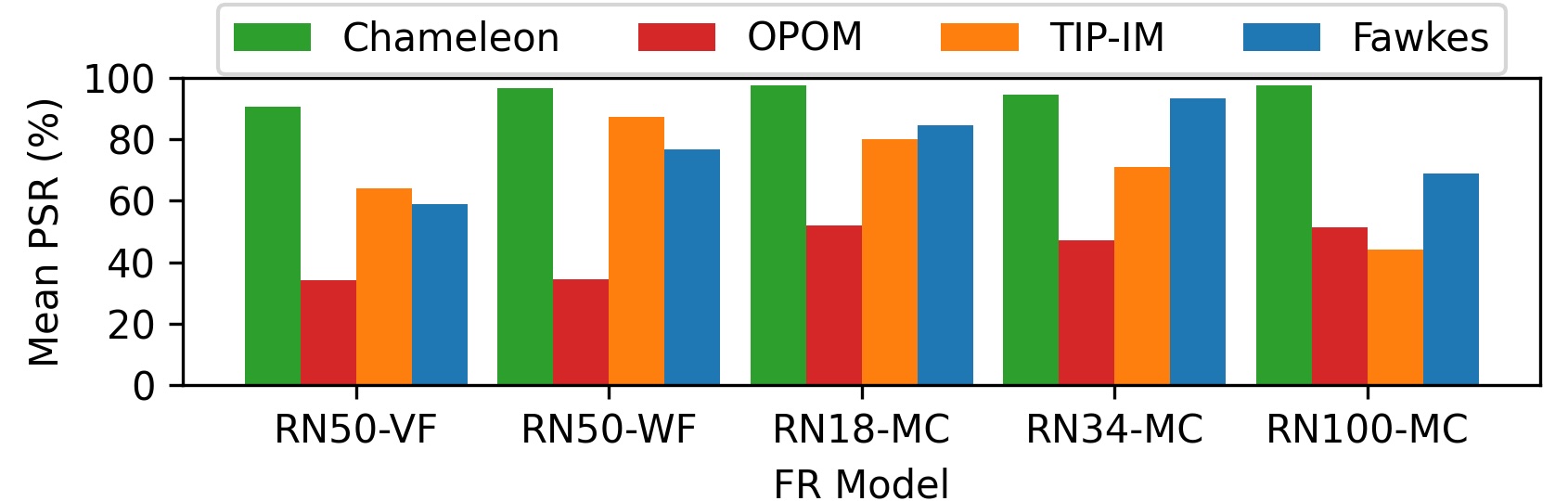}
		\caption{Dataset: FaceScrub}
		\label{fig:bar-facescrub}
	\end{subfigure}
	\begin{subfigure}{0.49\linewidth}
		\centering
		\includegraphics[width=\linewidth]{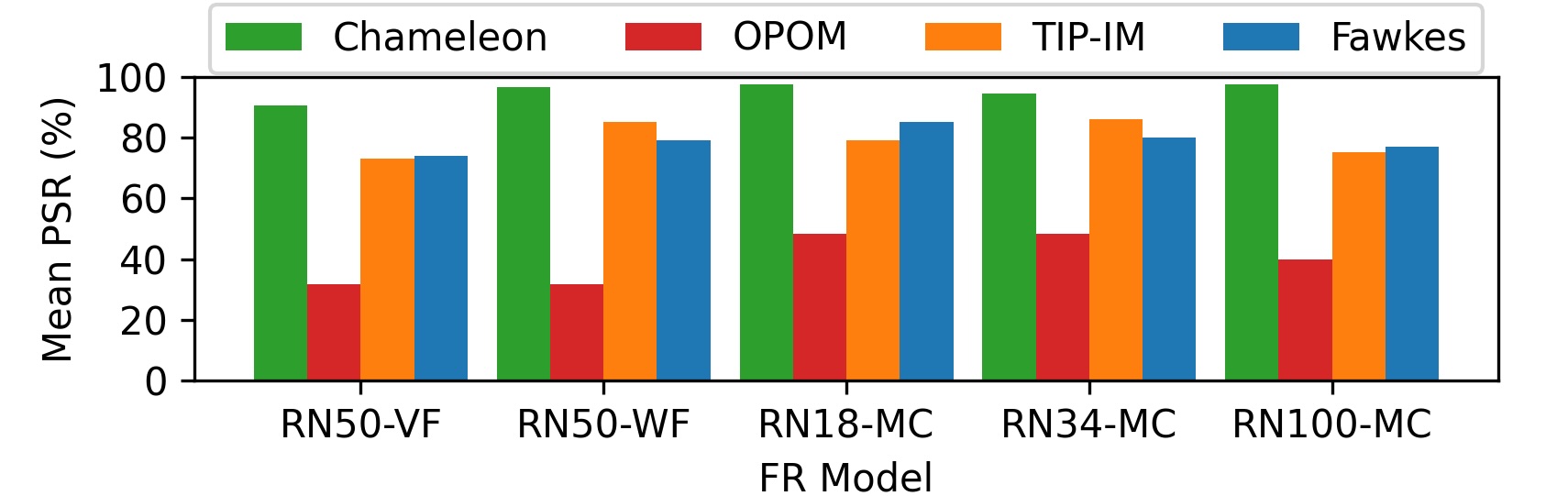}
		\caption{Dataset: LFW}
		\label{fig:bar-lfw}
	\end{subfigure}
	\caption{Protection effectiveness on two benchmarks using \scheme{}, OPOM~\cite{zhong2022opom}, TIP-IM~\cite{yang2021towards}, and Fawkes~\cite{shan2020fawkes} against FR models unknown to them.}
	\label{fig:bar}
\end{figure}

\scheme{} can protect against unknown FR models. 
We compare the cross-model protection of \scheme{} with OPOM~\cite{zhong2022opom}, the only anti-FR approach generating one privacy protection mask for each person as \scheme{}, and both TIP-IM~\cite{yang2021towards} and Fawkes~\cite{shan2020fawkes}, which conduct per-image optimization. For a fair comparison, all methods are set with the same perturbation budget. 
{Figure~\ref{fig:bar}} summarizes the results on both datasets. 
All five FR models are $\underline{\emph{unknown}}$ to both protection mechanisms. We make two observations. First, \scheme{} consistently outperforms OPOM by a large margin. We found that OPOM masks do not transfer well, especially considering the end-to-end ML pipeline. Its PSR can drop by $33\%$ because of those lossy operations.
Second, \scheme{} can be as competitive as those methods conducting per-image optimization. 
TIP-IM and Fawkes spend minutes to optimize the protection for \emph{each} image. Still, \scheme{} outperforms them and can be applied instantly. 

{Table~\ref{tab:psr}} reports the detailed PSR for each user \scheme{} achieves when the privacy intruder uses different FR models ($2$nd to $9$th columns). 
\begin{table}\scriptsize
	\centering
	\setlength\tabcolsep{2.7pt}
	\caption{Using the two-model team with high focal diversity ($2$nd and $3$rd columns), \scheme{} can offer privacy protection even when the privacy intruder uses FR models unseen during the mask generation process ($4$th to $9$th columns).}
	\label{tab:psr}
	\begin{tabular}{|l|cc|cccccc|cc|}
		\hline
		\multicolumn{1}{|c|}{\multirow{2}{*}{\textbf{Name}}} &
		\multicolumn{10}{c|}{\textbf{Protection Success Rate - PSR (\%)}} \\ \cline{2-11}
		\multicolumn{1}{|c|}{} &
		\multicolumn{1}{c|}{\textbf{\begin{tabular}[c]{@{}c@{}}EN-\\ MC\end{tabular}}} &
		\textbf{\begin{tabular}[c]{@{}c@{}}RN50-\\ GL\end{tabular}} &
		\multicolumn{1}{c|}{\textbf{\begin{tabular}[c]{@{}c@{}}RN50-\\ MC\end{tabular}}} &
		\multicolumn{1}{c|}{\textbf{\begin{tabular}[c]{@{}c@{}}RN50-\\ VF\end{tabular}}} &
		\multicolumn{1}{c|}{\textbf{\begin{tabular}[c]{@{}c@{}}RN50-\\ WF\end{tabular}}} &
		\multicolumn{1}{c|}{\textbf{\begin{tabular}[c]{@{}c@{}}RN18-\\ MC\end{tabular}}} &
		\multicolumn{1}{c|}{\textbf{\begin{tabular}[c]{@{}c@{}}RN34-\\ MC\end{tabular}}} &
		\textbf{\begin{tabular}[c]{@{}c@{}}RN100-\\ MC\end{tabular}} &
		\multicolumn{1}{c|}{\textbf{Mean}} & \textbf{Std}
		\\ \hline
		M. Baccarin &
		\multicolumn{1}{c|}{100.00} &
		100.00 &
		\multicolumn{1}{c|}{100.00} &
		\multicolumn{1}{c|}{72.73} &
		\multicolumn{1}{c|}{90.91} &
		\multicolumn{1}{c|}{100.00} &
		\multicolumn{1}{c|}{90.91} &
		100.00 &
		\multicolumn{1}{c|}{94.32} &
		9.64 \\ \hline
		B. Cooper &
		\multicolumn{1}{c|}{100.00} &
		100.00 &
		\multicolumn{1}{c|}{83.33} &
		\multicolumn{1}{c|}{100.00} &
		\multicolumn{1}{c|}{100.00} &
		\multicolumn{1}{c|}{100.00} &
		\multicolumn{1}{c|}{91.67} &
		91.67 &
		\multicolumn{1}{c|}{95.83} &
		6.30 \\ \hline
		A. Ferrera &
		\multicolumn{1}{c|}{100.00} &
		100.00 &
		\multicolumn{1}{c|}{93.75} &
		\multicolumn{1}{c|}{75.00} &
		\multicolumn{1}{c|}{100.00} &
		\multicolumn{1}{c|}{100.00} &
		\multicolumn{1}{c|}{93.75} &
		100.00 &
		\multicolumn{1}{c|}{95.31} &
		8.68 \\ \hline
		G. Butler &
		\multicolumn{1}{c|}{100.00} &
		100.00 &
		\multicolumn{1}{c|}{90.91} &
		\multicolumn{1}{c|}{100.00} &
		\multicolumn{1}{c|}{100.00} &
		\multicolumn{1}{c|}{90.91} &
		\multicolumn{1}{c|}{90.91} &
		100.00 &
		\multicolumn{1}{c|}{96.59} &
		4.70 \\ \hline
		E. Longoria &
		\multicolumn{1}{c|}{100.00} &
		100.00 &
		\multicolumn{1}{c|}{100.00} &
		\multicolumn{1}{c|}{100.00} &
		\multicolumn{1}{c|}{100.00} &
		\multicolumn{1}{c|}{100.00} &
		\multicolumn{1}{c|}{100.00} &
		100.00 &
		\multicolumn{1}{c|}{100.00} &
		0.00 \\ \hline
		M. Egan &
		\multicolumn{1}{c|}{100.00} &
		100.00 &
		\multicolumn{1}{c|}{100.00} &
		\multicolumn{1}{c|}{100.00} &
		\multicolumn{1}{c|}{100.00} &
		\multicolumn{1}{c|}{100.00} &
		\multicolumn{1}{c|}{100.00} &
		91.67 &
		\multicolumn{1}{c|}{98.96} &
		2.95 \\ \hline
		K. Cattrall &
		\multicolumn{1}{c|}{100.00} &
		100.00 &
		\multicolumn{1}{c|}{100.00} &
		\multicolumn{1}{c|}{84.62} &
		\multicolumn{1}{c|}{100.00} &
		\multicolumn{1}{c|}{100.00} &
		\multicolumn{1}{c|}{100.00} &
		100.00 &
		\multicolumn{1}{c|}{98.08} &
		5.44 \\ \hline
		A. Janney &
		\multicolumn{1}{c|}{100.00} &
		100.00 &
		\multicolumn{1}{c|}{91.67} &
		\multicolumn{1}{c|}{100.00} &
		\multicolumn{1}{c|}{91.67} &
		\multicolumn{1}{c|}{100.00} &
		\multicolumn{1}{c|}{100.00} &
		100.00 &
		\multicolumn{1}{c|}{97.92} &
		3.86 \\ \hline
		R. Downey &
		\multicolumn{1}{c|}{100.00} &
		100.00 &
		\multicolumn{1}{c|}{100.00} &
		\multicolumn{1}{c|}{72.73} &
		\multicolumn{1}{c|}{90.91} &
		\multicolumn{1}{c|}{100.00} &
		\multicolumn{1}{c|}{90.91} &
		100.00 &
		\multicolumn{1}{c|}{94.32} &
		9.64 \\ \hline
		S. Carell &
		\multicolumn{1}{c|}{100.00} &
		100.00 &
		\multicolumn{1}{c|}{84.62} &
		\multicolumn{1}{c|}{92.31} &
		\multicolumn{1}{c|}{92.31} &
		\multicolumn{1}{c|}{92.31} &
		\multicolumn{1}{c|}{76.92} &
		100.00 &
		\multicolumn{1}{c|}{92.31} &
		8.22 \\ \hhline{|=|=|=|=|=|=|=|=|=|=|=|}
		\multicolumn{1}{|r|}{\textbf{Mean}} &
		\multicolumn{1}{c|}{100.00} &
		100.00 &
		\multicolumn{1}{c|}{93.52} &
		\multicolumn{1}{c|}{90.65} &
		\multicolumn{1}{c|}{96.58} &
		\multicolumn{1}{c|}{97.41} &
		\multicolumn{1}{c|}{94.42} &
		97.42 &
		\multicolumn{2}{c|}{\multirow{2}{*}{}} \\ \cline{1-9}
		\multicolumn{1}{|r|}{\textbf{Std}} &
		\multicolumn{1}{c|}{0.00} &
		0.00 &
		\multicolumn{1}{c|}{6.40} &
		\multicolumn{1}{c|}{11.16} &
		\multicolumn{1}{c|}{4.43} &
		\multicolumn{1}{c|}{4.18} &
		\multicolumn{1}{c|}{7.41} &
		4.15 &
		\multicolumn{2}{c|}{} \\ \hline
	\end{tabular}
\end{table}
We make two observations. First, when the privacy intruder uses an FR model known to the generation process, a PSR of $100\%$ can be consistently achieved ($2$nd to $3$rd columns). Second, \scheme{} can protect against unknown FR models ($4$th to $9$th columns). Even when the FR model used by the intruder is trained on a different dataset (i.e., RN50-VF and RN50-WF) or with a different neural architecture (i.e., RN18-MC, RN34-MC, and RN100-MC) than any FR models known by \scheme{}, it still provides a PSR over $90.65\%$, meaning that facial images of a user are matched to gallery images of a different person. To demonstrate such a mismatch, in {Table~\ref{tab:matches}}, 
we show the probe images from two users and their most similar gallery images found by four unknown FR models with two settings: (i) the ``Unprotected" scenario where no one employs protection and (ii) the ``\scheme{}" scenario where the intruder scraped photos protected by our solution. Taking M. Baccarin as an example, the most similar gallery images found in the unprotected scenario belong to her. In contrast, when \scheme{} is used, the same probe image is misidentified, e.g., as L. Hartley by RN50-MC. 

\begin{table*}[t]\scriptsize
	\setlength\tabcolsep{0pt}
	\caption{The most similar gallery images on FaceScrub found by different FR models unknown to \scheme{}. }
	\label{tab:matches}
	\begin{tabular}{|c|cccccccc|}
		\hline
		\multirow{3}{*}{\textbf{\begin{tabular}[c]{@{}c@{}}Probe\\ Image\end{tabular}}} &
		\multicolumn{8}{c|}{\textbf{The Most Similar Gallery Image \& Its Identity Found by Unseen FR Models}} \\ \cline{2-9} 
		&
		\multicolumn{2}{c|}{\textbf{RN50-MC}} &
		\multicolumn{2}{c|}{\textbf{RN50-VF}} &
		\multicolumn{2}{c|}{\textbf{RN34-MC}} &
		\multicolumn{2}{c|}{\textbf{RN100-MC}} \\ \cline{2-9} 
		&
		\multicolumn{1}{c|}{\scriptsize\textbf{\tiny Unprotected}} &
		\multicolumn{1}{c|}{\scriptsize\textbf{\tiny \scheme{}}} &
		\multicolumn{1}{c|}{\scriptsize\textbf{\tiny Unprotected}} &
		\multicolumn{1}{c|}{\scriptsize\textbf{\tiny \scheme{}}} &
		\multicolumn{1}{c|}{\scriptsize\textbf{\tiny Unprotected}} &
		\multicolumn{1}{c|}{\scriptsize\textbf{\tiny \scheme{}}} &
		\multicolumn{1}{c|}{\scriptsize\textbf{\tiny Unprotected}} &
		{\scriptsize\textbf{\tiny \scheme{}}} \\ \hline
		\includegraphics[width=36.5px]{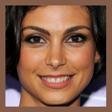} & \multicolumn{1}{c|}{\includegraphics[width=36.5px]{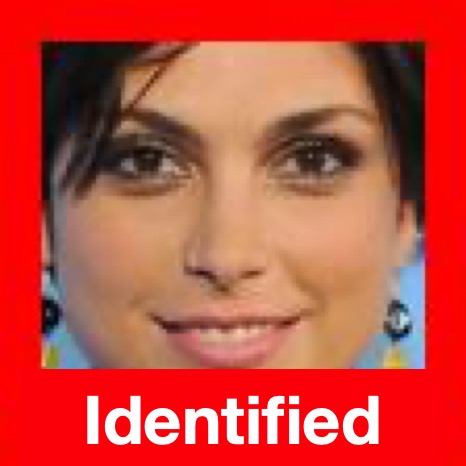}} & \multicolumn{1}{c|}{\includegraphics[width=36.5px]{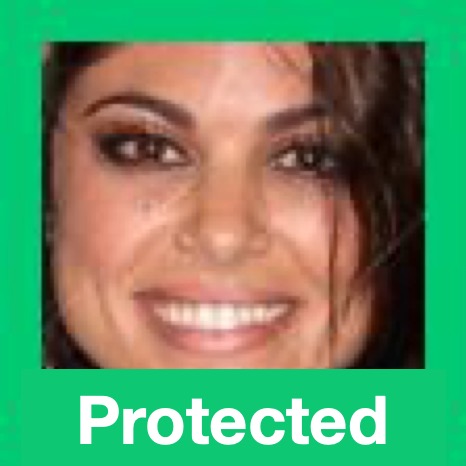}} & \multicolumn{1}{c|}{\includegraphics[width=36.5px]{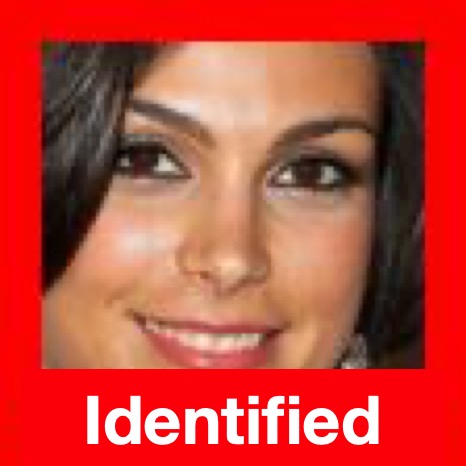}} & \multicolumn{1}{c|}{\includegraphics[width=36.5px]{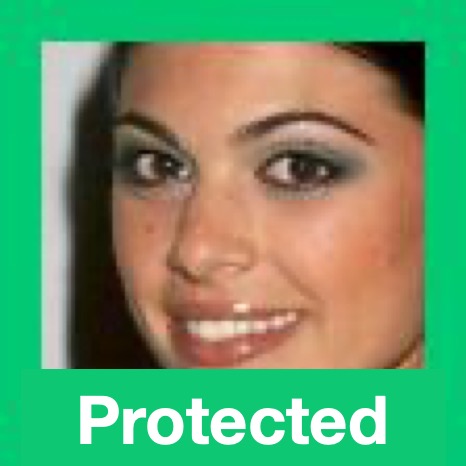}} & \multicolumn{1}{c|}{\includegraphics[width=36.5px]{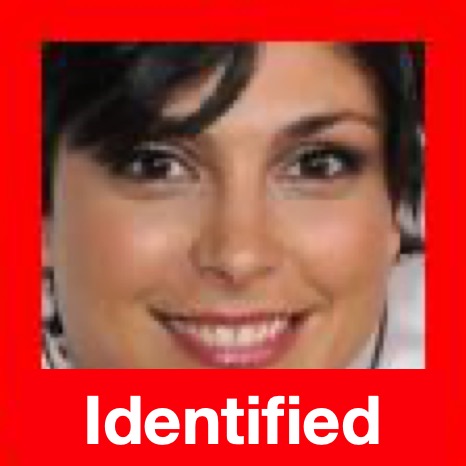}} & \multicolumn{1}{c|}{\includegraphics[width=36.5px]{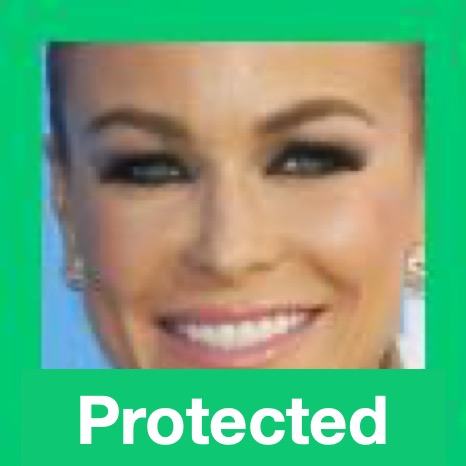}} & \multicolumn{1}{c|}{\includegraphics[width=36.5px]{figures/matches/Morena_Baccarin-c541c74c598c1481319d17ba0ff7dabd8893f1a7_Morena_Baccarin-0254e734383ea991e7b045c39758a0ed1c61d694.jpg}} & {\includegraphics[width=36.5px]{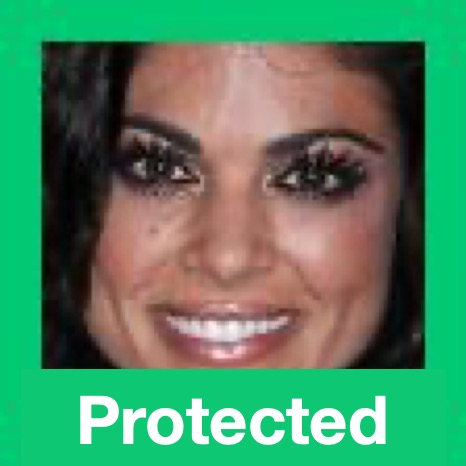}} \\
		{\bf\tiny M. Baccarin} & \multicolumn{1}{c|}{\bf\tiny\color{red} M. Baccarin} & \multicolumn{1}{c|}{\bf\tiny\color{teal} L. Hartley} & \multicolumn{1}{c|}{\bf\tiny\color{red} M. Baccarin} & \multicolumn{1}{c|}{\bf\tiny\color{teal} L. Hartley} & \multicolumn{1}{c|}{\bf\tiny\color{red} M. Baccarin} & \multicolumn{1}{c|}{\bf\tiny\color{teal} C. Electra} & \multicolumn{1}{c|}{\bf\tiny\color{red} M. Baccarin} & {\bf\tiny\color{teal} L. Hartley} \\ \hline
		\includegraphics[width=36.5px]{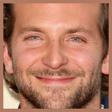} & \multicolumn{1}{c|}{\includegraphics[width=36.5px]{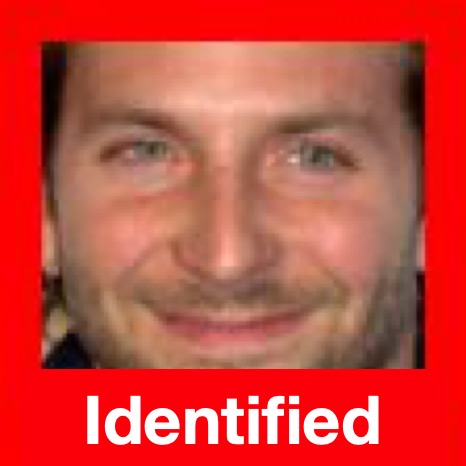}} & \multicolumn{1}{c|}{\includegraphics[width=36.5px]{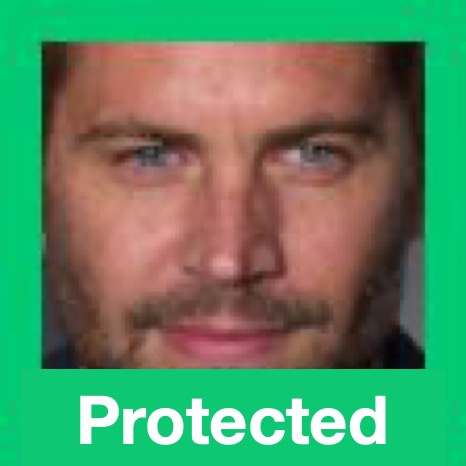}} & \multicolumn{1}{c|}{\includegraphics[width=36.5px]{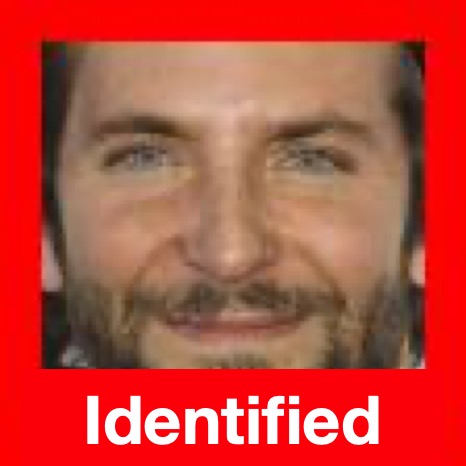}} & \multicolumn{1}{c|}{\includegraphics[width=36.5px]{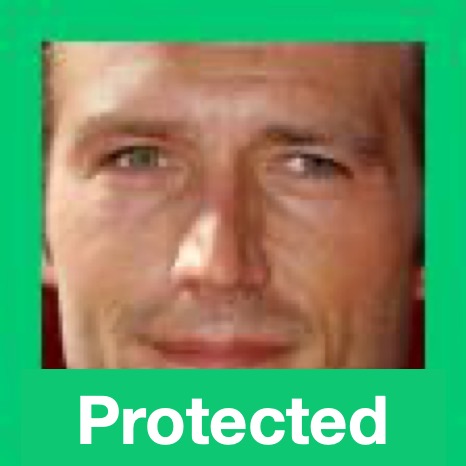}} & \multicolumn{1}{c|}{\includegraphics[width=36.5px]{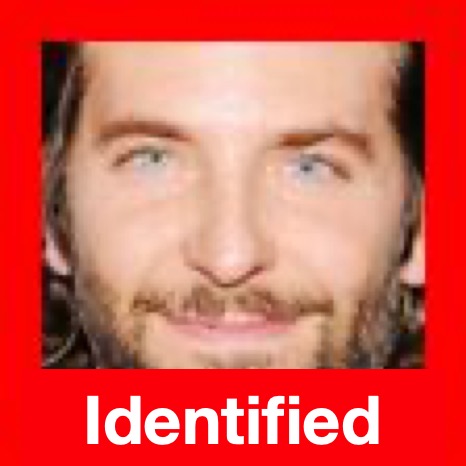}} & \multicolumn{1}{c|}{\includegraphics[width=36.5px]{figures/matches/Bradley_Cooper-d9f8000dc40d0317e7c46fde338f20506456152d_Michael_Vartan-4ec995fcfcde899a56ad94e626a0447296ba1ae4.jpg}} & \multicolumn{1}{c|}{\includegraphics[width=36.5px]{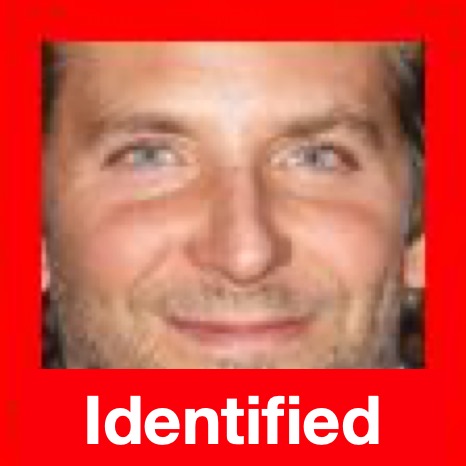}} & {\includegraphics[width=36.5px]{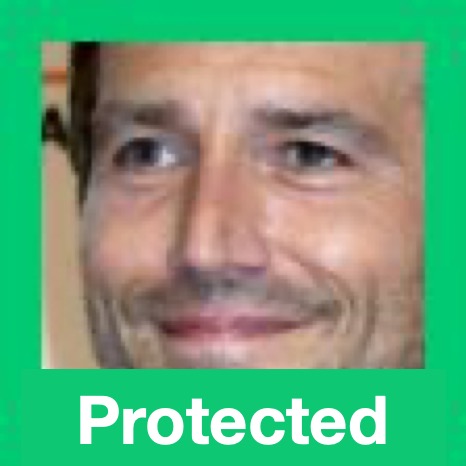}}  \\
		{\bf\tiny B. Cooper} & \multicolumn{1}{c|}{\bf\tiny\color{red} B. Cooper} & \multicolumn{1}{c|}{\bf\tiny\color{teal} P. Walker} & \multicolumn{1}{c|}{\bf\tiny\color{red} B. Cooper} & \multicolumn{1}{c|}{\bf\tiny\color{teal} M. Vartan}  & \multicolumn{1}{c|}{\bf\tiny\color{red} B. Cooper} & \multicolumn{1}{c|}{\bf\tiny\color{teal} M. Vartan} & \multicolumn{1}{c|}{\bf\tiny\color{red} B. Cooper} & {\bf\tiny\color{teal} M. Vartan} \\ \hline
	\end{tabular}
\end{table*}

\subsection{FR Service Usability}\label{sec:decrypt-results}
\scheme{} allows users to grant FR permission to trusted third parties by sharing their P3-Mask to de-obfuscate the protected image. 
There is no need to transmit and store an unprotected photo for internal use and a protected one for public visibility, doubling network and storage costs. {Table~\ref{tab:encrypt-decrypt}} shows the results in FR accuracy.
\begin{table*}[]\scriptsize
	\setlength\tabcolsep{0.6pt}
	\caption{The protection by \scheme{} can reduce FR accuracy against FR models (b), but the user can provide the P3-Mask used to protect her images as the key for the authorized third party to reverse the protection process (c), which leads to restored FR performance. However, an incorrect key cannot restore FR and may worsen it (d). }
	\label{tab:encrypt-decrypt}
	\begin{tabular}{|l|cccccccc|c|c|}
		\hline
		 \multicolumn{1}{|c|}{\multirow{3}{*}{\textbf{Scenario}}} & \multicolumn{10}{c|}{\textbf{Face Recognition Accuracy (\%)}}                                             \\ \cline{2-11}
		\multicolumn{1}{|c|}{}                                   & 
		\multicolumn{1}{c|}{\textbf{\begin{tabular}[c]{@{}c@{}}EN-\\ MC\end{tabular}}} & 
		\multicolumn{1}{c|}{\textbf{\begin{tabular}[c]{@{}c@{}}RN50-\\ GL\end{tabular}}} & 
		\multicolumn{1}{c|}{\textbf{\begin{tabular}[c]{@{}c@{}}RN50-\\ MC\end{tabular}}} & 
		\multicolumn{1}{c|}{\textbf{\begin{tabular}[c]{@{}c@{}}RN50-\\ VF\end{tabular}}} & 
		\multicolumn{1}{c|}{\textbf{\begin{tabular}[c]{@{}c@{}}RN50-\\ WF\end{tabular}}} &
		 \multicolumn{1}{c|}{\textbf{\begin{tabular}[c]{@{}c@{}}RN18-\\ MC\end{tabular}}} & 
		 \multicolumn{1}{c|}{\textbf{\begin{tabular}[c]{@{}c@{}}RN34-\\ MC\end{tabular}}} & 
		 \textbf{\begin{tabular}[c]{@{}c@{}}RN100-\\ MC\end{tabular}} &    
		  \textbf{Mean}                           &             \textbf{Std}                  \\ \hline
		\multicolumn{11}{|l|}{User: M. Baccarin}                    \\ \cline{1-11} 
		(a) No Protection                                           & \multicolumn{1}{c|}{100.00}         & \multicolumn{1}{c|}{100.00}           & \multicolumn{1}{c|}{100.00}           & \multicolumn{1}{c|}{100.00}           & \multicolumn{1}{c|}{100.00}           & \multicolumn{1}{c|}{100.00}           & \multicolumn{1}{c|}{100.00}           & 100.00            & 100.00                         & 0.00                          \\ \cline{1-11} 
		 (b) Protected                           & \multicolumn{1}{c|}{0.00}           & \multicolumn{1}{c|}{0.00}             & \multicolumn{1}{c|}{9.09}             & \multicolumn{1}{c|}{18.18}            & \multicolumn{1}{c|}{9.09}             & \multicolumn{1}{c|}{9.09}             & \multicolumn{1}{c|}{0.00}             & 9.09              & 6.82                           & 6.43                          \\ \cline{1-11} 
		 (c) Correctly Unmasked                           & \multicolumn{1}{c|}{100.00}         & \multicolumn{1}{c|}{100.00}           & \multicolumn{1}{c|}{100.00}           & \multicolumn{1}{c|}{100.00}           & \multicolumn{1}{c|}{100.00}           & \multicolumn{1}{c|}{100.00}           & \multicolumn{1}{c|}{100.00}           & 100.00            & 100.00                         & 0.00                          \\\cline{1-11} 
		 (d) Incorrectly Unmasked                          & \multicolumn{1}{c|}{0.00}           & \multicolumn{1}{c|}{0.00}             & \multicolumn{1}{c|}{0.00}             & \multicolumn{1}{c|}{0.00}             & \multicolumn{1}{c|}{9.09}             & \multicolumn{1}{c|}{0.00}             & \multicolumn{1}{c|}{0.00}             & 0.00              & 1.14                           & 3.21                          \\ \hhline{|=|=|=|=|=|=|=|=|=|=|=|}
		 \multicolumn{11}{|l|}{User: B. Cooper}                    \\ \cline{1-11} 
		(a) No Protection                                           & \multicolumn{1}{c|}{100.00}         & \multicolumn{1}{c|}{100.00}           & \multicolumn{1}{c|}{100.00}           & \multicolumn{1}{c|}{100.00}           & \multicolumn{1}{c|}{100.00}           & \multicolumn{1}{c|}{100.00}           & \multicolumn{1}{c|}{100.00}           & 100.00            & 100.00                         & 0.00                          \\ \cline{1-11} 
		(b) Protected                           & \multicolumn{1}{c|}{0.00}           & \multicolumn{1}{c|}{0.00}             & \multicolumn{1}{c|}{16.67}            & \multicolumn{1}{c|}{0.00}             & \multicolumn{1}{c|}{0.00}             & \multicolumn{1}{c|}{0.00}             & \multicolumn{1}{c|}{8.33}             & 8.33           &    4.17                           & 6.30                          \\ \cline{1-11} 
	 (c) Correctly Unmasked                           & \multicolumn{1}{c|}{100.00}         & \multicolumn{1}{c|}{100.00}           & \multicolumn{1}{c|}{100.00}           & \multicolumn{1}{c|}{100.00}           & \multicolumn{1}{c|}{100.00}           & \multicolumn{1}{c|}{100.00}           & \multicolumn{1}{c|}{100.00}           & 100.00            & 100.00                         & 0.00                          \\ \cline{1-11} 
		 (d) Incorrectly Unmasked                          & \multicolumn{1}{c|}{0.00}           & \multicolumn{1}{c|}{0.00}             & \multicolumn{1}{c|}{0.00}             & \multicolumn{1}{c|}{0.00}             & \multicolumn{1}{c|}{0.00}             & \multicolumn{1}{c|}{0.00}             & \multicolumn{1}{c|}{0.00}             & 0.00              & 0.00                           & 0.00                          \\ \hline
	\end{tabular}
\end{table*}
First, using the correct mask (i.e., the one used to protect the images) to reverse the protection can restore FR accuracy. Taking M. Baccarin as an example, the FR accuracy increases from $6.82\%$ before unmasking (b) to $100\%$ after unmasking (c). Second, using an incorrect mask  for unmasking cannot restore the FR accuracy and can lead to even worse performance. It drops from $6.82\%$ to $1.14\%$ after unmasking (d).

\subsection{Protection Cost Analysis}\label{sec:eval-cost}
\begin{figure}[t]
	\centering
	\includegraphics[width=0.9\linewidth]{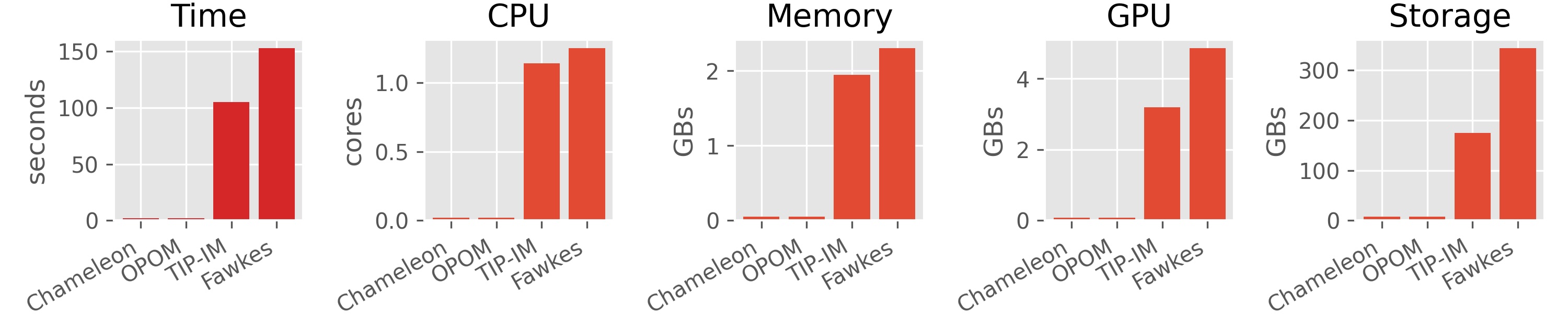}
	\caption{\scheme{} offers much better privacy protection than OPOM without sacrificing speed and resource usages. It can protect instantly and require negligible compute resources as OPOM, while TIP-IM and Fawkes are slow and costly.}
	\label{fig:cost}
\end{figure}

\textbf{Speed and Resources.} {Figure~\ref{fig:cost}} compares \scheme{} with OPOM, TIP-IM, and Fawkes regarding the protection time per face, compute, and storage costs. 
\scheme{} and OPOM produce masks that are applicable to any facial image of the same user. The protection can be completed in $0.0076$ seconds. Note that \scheme{} is significantly more effective than OPOM, as described in Figure~\ref{fig:bar}. Instead, TIP-IM and Fawkes require per-image optimization and take $105.12$ seconds to complete the protection with GPUs for acceleration and storage for  FR models. \scheme{} only needs to conduct simple arithmetic operations. Only the P3-Mask needs to be stored on the user device. Hence, it can be deployed with instant protection even on an edge device with weak computing power.

\textbf{Image Quality.}  \scheme{} can well preserve image quality. {Table~\ref{tab:image-quality-visual}} provides two examples for four users, contrasting the facial images with no protection ($3$rd and $5$th columns) with the ones with their facial signature removed by \scheme{} ($4$th and $6$th columns). \begin{table}\scriptsize
	\begin{minipage}[t]{.65\textwidth}
		\centering\scriptsize
		\caption{The P3-Mask can be applied to any facial images of the same person while preserving image quality.}
		\setlength\tabcolsep{1pt}
		\label{tab:image-quality-visual}
		\begin{tabular}{|l|c|c|c|c|c|}
			\hline
			\multicolumn{1}{|c|}{\multirow{3}{*}{\textbf{User}}} & \multirow{3}{*}{\textbf{Mask}} & \multicolumn{2}{|c|}{\textbf{Sample Image 1}} & \multicolumn{2}{|c|}{\textbf{Sample Image 2}} \\ \cline{3-6}
			\multicolumn{1}{|c|}{} &  & \textbf{\tiny\begin{tabular}[c]{@{}c@{}}No\\ Protection\end{tabular}} & \textbf{\tiny Protected} & \textbf{\tiny\begin{tabular}[c]{@{}c@{}}No\\ Protection\end{tabular}} & \textbf{ \tiny Protected} \\ \hline
			\begin{tabular}[c]{@{}l@{}}M.\\ Baccarin\end{tabular} & \raisebox{-.45\height}{\includegraphics[width=35px]{figures/masks/Morena_Baccarin.jpg}} & \raisebox{-.45\height}{\includegraphics[width=35px]{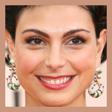}} & \raisebox{-.45\height}{\includegraphics[width=35px]{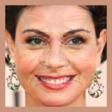}} & \raisebox{-.45\height}{\includegraphics[width=35px]{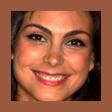}} & \raisebox{-.45\height}{\includegraphics[width=35px]{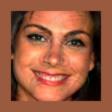}} \\ \hline
			\begin{tabular}[c]{@{}l@{}}B.\\ Cooper\end{tabular} & \raisebox{-.45\height}{\includegraphics[width=35px]{figures/masks/Bradley_Cooper.jpg}} & \raisebox{-.45\height}{\includegraphics[width=35px]{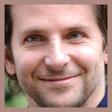}} & \raisebox{-.45\height}{\includegraphics[width=35px]{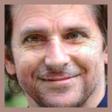}} & \raisebox{-.45\height}{\includegraphics[width=35px]{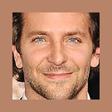}} & \raisebox{-.45\height}{\includegraphics[width=35px]{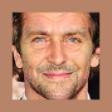}} \\ \hline
			\begin{tabular}[c]{@{}l@{}}E.\\ Longoria\end{tabular}  & \raisebox{-.45\height}{\includegraphics[width=35px]{figures/masks/Eva_Longoria.jpg}} & \raisebox{-.45\height}{\includegraphics[width=35px]{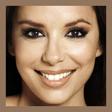}} & \raisebox{-.45\height}{\includegraphics[width=35px]{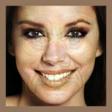}} & \raisebox{-.45\height}{\includegraphics[width=35px]{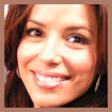}} & \raisebox{-.45\height}{\includegraphics[width=35px]{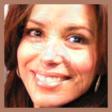}} \\ \hline
			\begin{tabular}[c]{@{}l@{}}G.\\ Image\end{tabular}  & \raisebox{-.45\height}{\includegraphics[width=35px]{figures/masks/Gerard_Butler.jpg}} & \raisebox{-.45\height}{\includegraphics[width=35px]{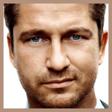}} & \raisebox{-.45\height}{\includegraphics[width=35px]{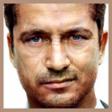}} & \raisebox{-.45\height}{\includegraphics[width=35px]{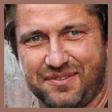}} & \raisebox{-.45\height}{\includegraphics[width=35px]{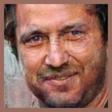}} \\ \hline
		\end{tabular}
	\end{minipage}
	\hfill
	\begin{minipage}[t]{0.335\textwidth}\scriptsize
		\centering
		\caption{\scheme{} offers much better protection while maintaining image quality.}\label{tab:quant-vis}
		\begin{tabular}{|l|c|}
			\hline
			& \textbf{ Example} \\ \hline
			\begin{tabular}[c]{@{}l@{}}Original\\ {\textbf{SSIM}: 1.0000}\end{tabular}   &  \raisebox{-.5\height}{\includegraphics[width=0.255\linewidth]{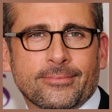}}                                                               \\ \hline
			 \begin{tabular}[c]{@{}l@{}}Chameleon\\ {\textbf{SSIM}: 0.9493}\end{tabular}    &   \raisebox{-.45\height}{\includegraphics[width=0.255\linewidth]{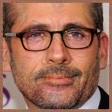}    }                                                                 \\ \hline
			\begin{tabular}[c]{@{}l@{}}OPOM\\ {\textbf{SSIM}: 0.8839}\end{tabular}           &     \raisebox{-.45\height}{\includegraphics[width=0.255\linewidth]{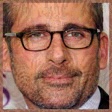}   }                                                                \\ \hline
			 \begin{tabular}[c]{@{}l@{}}TIP-IM\\ {\textbf{SSIM}: 0.8850}\end{tabular}      &   \raisebox{-.45\height}{\includegraphics[width=0.255\linewidth]{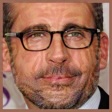}   }                                                                    \\ \hline
			 \begin{tabular}[c]{@{}l@{}}Fawkes\\ {\textbf{SSIM}: 0.9612}  \end{tabular}      &  \raisebox{-.45\height}{\includegraphics[width=0.255\linewidth]{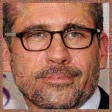}   }                                                                       \\ \hline
		\end{tabular}
	\end{minipage}
\end{table}
They capture different variations of facial images: M. Baccarin’s images have different lighting conditions, B. Cooper’s have different face sizes, E. Longoria’s have different postures, and G. Bulter’s have different expressions. The protected images are visually similar to the unprotected counterparts, but they will not be matched to the clean images of the corresponding person.  
Indeed, \scheme{} can generate higher-quality images than existing methods. As reported in {Table~\ref{tab:quant-vis}}, \scheme{} not only offers better protection (Figure~\ref{fig:bar}) with a similar cost (Figure~\ref{fig:cost}) as OPOM but also achieves much better image quality measured in SSIM ($0.9493$ vs $0.8839$).

\subsection{Focal Diversity-based Teaming}\label{sec:eval-time}
\scheme{} automatically selects a high-quality team for deployment with a specified budget. {Table~\ref{tab:focal-prot-tab}} compares the detailed PSR using the most diverse and the least diverse teams with (a) two and (b) three models.
\begin{table*}[b]\scriptsize
	\setlength\tabcolsep{0pt}
	\centering
	\caption{The most diverse and the least diverse teams with (a) two or (b) three models identified by our focal diversity-based teaming method. The most diverse teams offer significantly better protection in terms of protection success rates.}
	\label{tab:focal-prot-tab}
	\begin{tabular}{|lcccccccccc|}
		\hline
		\multicolumn{1}{|c|}{\multirow{2}{*}{\textbf{Protection Team}}}  & \multicolumn{10}{c|}{\textbf{Protection Success Rate - PSR (\%)}}                  \\ \cline{2-11}
		\multicolumn{1}{|c|}{}                                           & 
		\multicolumn{1}{c|}{\textbf{\begin{tabular}[c]{@{}c@{}}EN-\\ MC\end{tabular}}} & 
		\multicolumn{1}{c|}{\textbf{\begin{tabular}[c]{@{}c@{}}RN50-\\ GL\end{tabular}}} & 
		\multicolumn{1}{c|}{\textbf{\begin{tabular}[c]{@{}c@{}}RN50-\\ MC\end{tabular}}} & 
		\multicolumn{1}{c|}{\textbf{\begin{tabular}[c]{@{}c@{}}RN50-\\ VF\end{tabular}}} & 
		\multicolumn{1}{c|}{\textbf{\begin{tabular}[c]{@{}c@{}}RN50-\\ WF\end{tabular}}} & 
		\multicolumn{1}{c|}{\textbf{\begin{tabular}[c]{@{}c@{}}RN18-\\ MC\end{tabular}}} &
		 \multicolumn{1}{c|}{\textbf{\begin{tabular}[c]{@{}c@{}}RN34-\\ MC\end{tabular}}} & 
		 \multicolumn{1}{c|}{\textbf{\begin{tabular}[c]{@{}c@{}}RN100-\\ MC\end{tabular}}} & 
		 \multicolumn{1}{c|}{\textbf{Mean}}                               &                          \textbf{Std}     \\ \hline
		\multicolumn{11}{|l|}{(a) Two-model Teams}                                                                                                                                                                                                                                                                                                                                                                                                                                           \\ \hline
		\multicolumn{1}{|l|}{\begin{tabular}[c]{@{}l@{}}\begin{tabular}[c]{@{}l@{}}Most Diverse: \\ {\tiny (EN-MC, RN50-GL)}\end{tabular}\end{tabular} }             & \multicolumn{1}{c|}{100.00}         & \multicolumn{1}{c|}{100.00}           & \multicolumn{1}{c|}{93.52}            & \multicolumn{1}{c|}{90.65}            & \multicolumn{1}{c|}{96.58}            & \multicolumn{1}{c|}{97.41}            & \multicolumn{1}{c|}{94.42}            & \multicolumn{1}{c|}{97.42}             & \multicolumn{1}{c|}{\textbf{96.25}}                 & \textbf{3.23}                 \\ \hline
		\multicolumn{1}{|l|}{\begin{tabular}[c]{@{}l@{}}\begin{tabular}[c]{@{}l@{}}Least Diverse:\\{\tiny (RN18-MC, RN34-MC)}\end{tabular}\end{tabular}}          & \multicolumn{1}{c|}{11.38}           & \multicolumn{1}{c|}{60.16}            & \multicolumn{1}{c|}{100.00}           & \multicolumn{1}{c|}{55.28}            & \multicolumn{1}{c|}{73.17}            & \multicolumn{1}{c|}{100.00}           & \multicolumn{1}{c|}{100.00}           & \multicolumn{1}{c|}{80.49}             & \multicolumn{1}{c|}{72.56}                          & 28.54                         \\ \hhline{|=|=|=|=|=|=|=|=|=|=|=|}
		\multicolumn{11}{|l|}{(b) Three-model Teams}                                                                                                                                                                                                                                                                                                                                                                                                                                         \\ \hline
		\multicolumn{1}{|l|}{\begin{tabular}[c]{@{}l@{}}\begin{tabular}[c]{@{}l@{}}Most Diverse:\\{\tiny (EN-MC, RN50-GL, RN50-WF)}\end{tabular}\end{tabular}}    & \multicolumn{1}{c|}{100.00}         & \multicolumn{1}{c|}{100.00}           & \multicolumn{1}{c|}{100.00}           & \multicolumn{1}{c|}{100.00}           & \multicolumn{1}{c|}{92.68}            & \multicolumn{1}{c|}{93.50}            & \multicolumn{1}{c|}{100.00}           & \multicolumn{1}{c|}{100.00}            & \multicolumn{1}{c|}{\textbf{98.27}}                 & \textbf{3.00}                 \\ \hline
		\multicolumn{1}{|l|}{\begin{tabular}[c]{@{}l@{}}\begin{tabular}[c]{@{}l@{}}Least Diverse:\\{\tiny (RN18-MC, RN34-MC, RN50-MC)}\end{tabular}\end{tabular}} & \multicolumn{1}{c|}{2.44}           & \multicolumn{1}{c|}{78.86}            & \multicolumn{1}{c|}{100.00}           & \multicolumn{1}{c|}{52.03}            & \multicolumn{1}{c|}{84.55}            & \multicolumn{1}{c|}{100.00}           & \multicolumn{1}{c|}{100.00}           & \multicolumn{1}{c|}{93.50}             & \multicolumn{1}{c|}{76.42}                          & 31.82                         \\ \hline
	\end{tabular}
\end{table*}
In both cases, the most diverse team outperforms the least diverse one, which is only effective when the privacy intruder uses the FR model known in the team. The selection of high-quality teams is non-trivial because we observe that composing a team of FR models with the highest FR accuracy is not always a good option, and a team of models having different neural architectures is not always the top priority~\cite{chow2021robust}.

In {Figure~\ref{fig:unknown-algo}}, we further show that the P3-Mask generated from a carefully-chosen team can protect against FR models of unknown algorithms. Specifically, we consider the most and the least diverse three-model teams in Table~\ref{tab:focal-prot-tab}(b). Even though both teams include only FR models based on ArcFace~\cite{deng2019arcface}, the most diverse one can effectively protect against FR models based on FaceNet~\cite{schroff2015facenet} or MagFace~\cite{meng2021magface}. The protection effectiveness can be further strengthened by incorporating more FR models into the collection.

\begin{table}[t]\scriptsize
	\begin{minipage}[t]{.43\textwidth}
		\centering
		\includegraphics[width=0.95\linewidth]{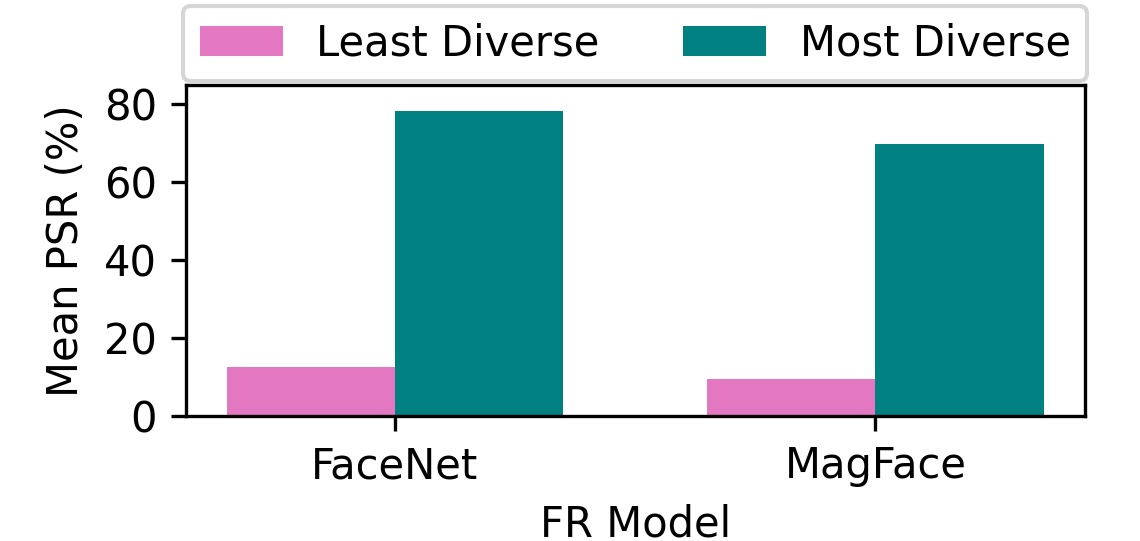}\vspace{0.7em}
		\captionof{figure}{The most diverse teams allow \scheme{} to be effective against FR models of unknown algorithms.}
		\label{fig:unknown-algo}
	\end{minipage}
	\hfill
	\begin{minipage}[t]{0.55\textwidth}\scriptsize
		\centering
		\includegraphics[width=0.95\linewidth]{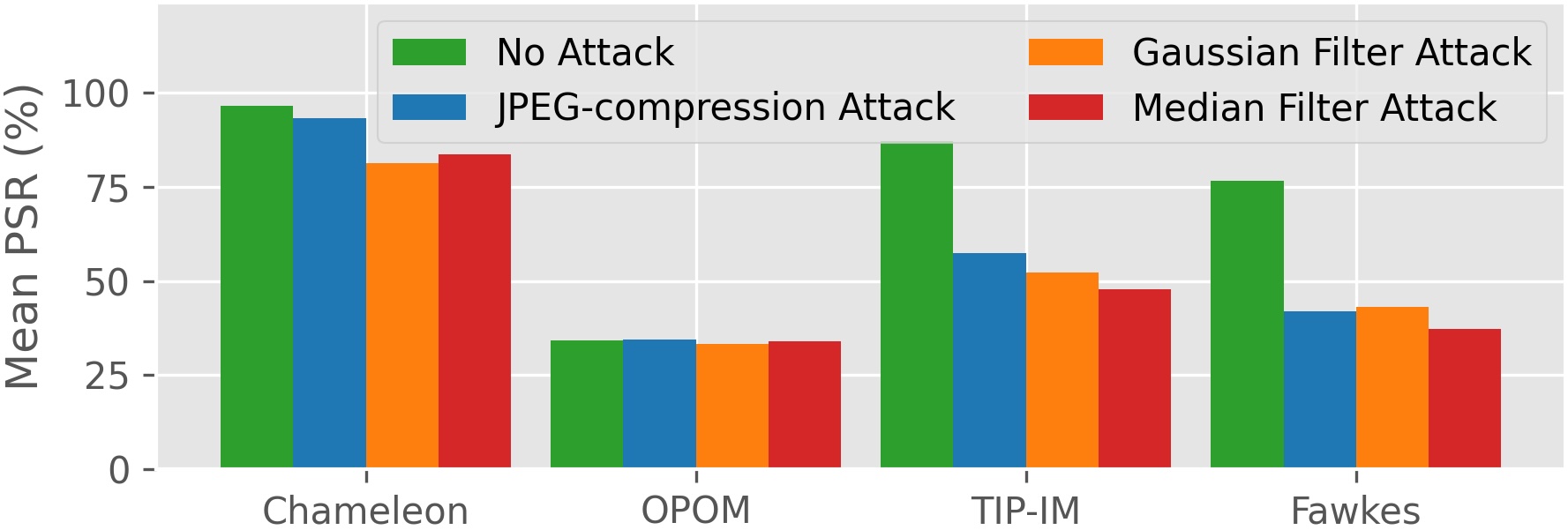}\vspace{0.7em}
		\captionof{figure}{The PSR of \scheme{} against adaptive intruders using different strategies to wash out the patterns in  protected images.}
		\label{fig:adaptive}
	\end{minipage}
\end{table}
\subsection{\scheme{} Against Adaptive Adversaries}
An adaptive adversary may run \scheme{} to produce the P3-Mask for each person and de-obfuscate their photos. When probe images need to be identified, they will be matched against a ``clean" face database. However, it is infeasible, as we show in Section~\ref{sec:decrypt-results}, that the FR accuracy can only be restored when the masks used for obfuscation and de-obfuscation are identical. It is impossible to reproduce the same P3-Mask used by the user because it requires the same set of clean images and random states.

Alternatively, an adaptive adversary may wash out patterns introduced by \scheme{} before performing FR. We consider three popular lightweight and task-agnostic methods studied in the adversarial example domain~\cite{guo2017countering} in {Figure~\ref{fig:adaptive}}, including (a) JPEG compression~\cite{das2018shield}, (b) Gaussian Filter, and (d) Median Filter~\cite{xu2018feature}. The results show that their effectiveness is limited on \scheme{}. The consideration of the end-to-end ML pipeline increases the robustness of P3-Mask.

\section{Conclusions}
We have presented \scheme{} against unauthorized FR. \scheme{} generates a P3-Mask for each user with cross-image and perceptibility optimizations to offer (i) instant and lightweight protection on any facial images of the same user and (ii) preservation of image quality. Also, we have shown that P3-Mask enables cost-effective de-obfuscation for authorizing FR services. Such an authorization process can only be conducted by the one with the identical mask used for protection. \scheme{} also features a focal diversity-optimized teaming method to select a high-quality FR team to generate P3-Mask with strong robustness against unknown FR models. 

\section*{Acknowledgments}
This research is partially sponsored by the NSF CISE grants 2302720, 2312758, 2038029, an IBM faculty award, and a grant from CISCO Edge AI program. It is part of the PhD dissertation of the first author, who graduated from Georgia Tech in Spring 2023. The first author acknowledges the support of the IBM PhD Fellowship in 2022-2023 and the support from the HKU-CS Start-up Fund.

\bibliographystyle{splncs04}
\bibliography{references}

\end{document}